# Distributed Planning in Hierarchical Factored MDPs

Carlos Guestrin
Computer Science Dept
Stanford University
*guestrin@cs.stanford.edu*

Geoffrey Gordon
Computer Science Dept
Carnegie Mellon University
*ggordon@cs.cmu.edu*

We present a principled and efficient planning algorithm for collaborative multiagent dynamical systems. All computation, during both the planning and the execution phases, is distributed among the agents; each agent only needs to model and plan for a small part of the system. Each of these local subsystems is small, but once they are combined they can represent an exponentially larger problem. The subsystems are connected through a subsystem hierarchy. Coordination and communication between the agents is not imposed, but derived directly from the structure of this hierarchy. A globally consistent plan is achieved by a message passing algorithm, where messages correspond to natural local reward functions and are computed by local linear programs; another message passing algorithm allows us to execute the resulting policy. When two portions of the hierarchy share the same structure, our algorithm can reuse plans and messages to speed up computation.

## 1 Introduction

Many interesting planning problems have a very large number of states and actions, described as the cross product of a small number of state variables and action variables. Even very fast exact algorithms cannot solve such large planning problems in a reasonable amount of time. Fortunately, in many such cases we can group the variables in these planning problems into subsystems that interact in a simple manner.

As argued by Herbert Simon [20] in "Architecture of Complexity", many complex systems have a "nearly decomposable, hierarchical structure", with the subsystems interacting only weakly between themselves. In this paper, we represent planning problems using a hierarchical decomposition into local subsystems. (In multiagent planning problems, each agent will usually implement one or more local subsystems.) Although each subsystem is small, once these subsystems are combined we can represent an *exponentially* larger problem.

The advantage of constructing such a grouping is that we can hope to plan for each subsystem separately. Coordinating many local planners to form a successful global plan requires careful attention to communication between planners: if two local plans make contradictory assumptions, global success is unlikely.

In this paper, we describe an algorithm for coordinating many local Markov decision process (MDP) planners to form a global plan. The algorithm is relatively simple: at each stage we solve a small number of small linear programs to determine messages that each planner will pass to its neighbors, and based on these messages the local planners revise their plans and send new messages until the plans stop changing. The messages have an appealing interpretation: they are rewards or penalties for taking actions that benefit or hurt their neighbors, and statistics about plans that the agents are considering.

Our hierarchical decomposition offers another significant feature: *reuse*. When two subsystems are equivalent (i.e., are instances of the same class), plans devised for one subsystem can be reused in the other. Furthermore, in many occasions, larger parts of the system may be equivalent. In these cases, we can not only reuse plans, but also messages.

The individual local planners can run any planning algorithm they desire, so long as they can extract a particular set of state visitation frequencies from their plans. Of course, suboptimal planning from local planners will tend to reduce the quality of the global plan.

Our algorithm does not necessarily converge to an optimal plan. However, it is guaranteed to be the same as the plan produced by a particular centralized planning algorithm (approximate linear programming as in [19], with a particular basis).

## 2 Related work

Many researchers have examined the idea of dividing a planning problem into simpler pieces in order to solve it faster. There are two common ways to split a problem into simpler pieces, which we will call serial decomposition and parallel decomposition. Our planning algorithm is significant because it handles both serial and parallel decompositions, and it provides more opportunities for abstraction than other parallel-decomposition planners. Additionally, it is fully distributed: at no time is there a global combination step requiring knowledge of all subproblems simultaneously. No previous algorithm combines these qualities.

In a *serial decomposition*, one subproblem is active at any given time. The overall state consists of an indicator of which subproblem is active along with that subproblem's state. Subproblems interact at their borders, which are the states where we can enter or leave them. For exam-ple, imagine a robot navigating in a building with multiple rooms connected by doorways: fixing the value of the doorway states decouples the rooms from each other and lets us solve each room separately. In this type of decomposition, the combined state space is the union of the subproblem state spaces, and so the total size of all of the subproblems is approximately equal to the size of the combined problem.

Serial decomposition planners in the literature include Kushner and Chen's algorithm [12] and Dean and Lin's algorithm [6], as well as a variety of hierarchical planning algorithms. Kushner and Chen were the first to apply Dantzig-Wolfe decomposition to Markov decision processes, while Dean and Lin combined decomposition with state abstraction. Hierarchical planning algorithms include MAXQ [7], hierarchies of abstract machines [16], and planning with macro-operators [22, 9].

By contrast, in a *parallel decomposition*, multiple subproblems can be active at the same time, and the combined state space is the cross product of the subproblem state spaces. The size of the combined problem is therefore exponential rather than linear in the number of subproblems, which means that a parallel decomposition can potentially save us much more computation than a serial one. For an example of a parallel decomposition, suppose there are multiple robots in our building, interacting through a common resource constraint such as limited fuel or through a common goal such as lifting a box which is too heavy for one robot to lift alone. A subproblem of this task might be to plan a path for one robot using only a compact summary of the plans for the other robots.

Parallel decomposition planners in the literature include Singh and Cohn's [21] and Meuleau et al.'s [15] algorithms. Singh and Cohn's planner builds the combined state space explicitly, using subproblem solutions to initialize the global search. So, while it may require fewer planning iterations than naive global planning, it is limited by having to enumerate an exponentially large set. Meuleau et al.'s planner is designed for parallel decompositions in which the only coupling is through global resource constraints. More complicated interactions such as conjunctive goals or shared state variables are beyond its scope.

Our planner works by representing a planning problem as a linear program [14], substituting in a compact approximate representation of the solution [19], and transforming and decomposing the LP so that it can be solved by a distributed network of planners. One of this paper's contributions is the method for transformation and decomposition.

Our transformation is based on the factorized planning algorithm of Guestrin, Koller, and Parr [8]. Their algorithm uses a central planner, but allows distributed execution of plans. We extend that result to allow planning to be distributed as well, while still guaranteeing that we reach the same solution. That means that our algorithm allows for truly decentralized multiagent planning and execution: each agent can run its own local planner and compute messages locally, and doesn't need to know the global state of the world or the actions of unrelated agents.

The transformation produces a sparse linear program, to which we apply a nested Benders decomposition [1]. (Or, dually, a Dantzig-Wolfe decomposition [3].) As mentioned above, other authors have used this decomposition for planning; but, our method is the first to handle parallel decompositions of planning problems.

Another contribution of our new hierarchical representation and planning algorithm over the algorithm of Guestrin et al. [8] is reuse. As we describe in Sec. 9, our approach can reuse plans and messages for parts of the hierarchy that share the same structure.

## 3 Markov Decision Processes

The *Markov Decision Process (MDP)* framework formalizes the problem where agents are acting in a dynamic environment, and are jointly trying to maximize their expected long-term return. An MDP is defined as a 4-tuple $(\mathbf{X}, \mathbf{A}, R, P)$ where: $\mathbf{X}$ is a finite set of $N = |\mathbf{X}|$ states; $\mathbf{A}$ is a set of actions; $R$ is a *reward function* $R : \mathbf{X} \times \mathbf{A} \mapsto \mathbb{R}$ such that $R(\mathbf{x}, \mathbf{a})$ represents the reward obtained in state $\mathbf{x}$ after taking action $\mathbf{a}$; and $P$ is a *Markovian transition model* where $P(\mathbf{x}' \mid \mathbf{x}, \mathbf{a})$ represents the probability of going from state $\mathbf{x}$ to state $\mathbf{x}'$ with action $\mathbf{a}$. We will write $R_\mathbf{a}$ to mean the vector of rewards associated with action $\mathbf{a}$ (with one entry for each state), and we will write $P_\mathbf{a}$ to mean the transition matrix associated with action $\mathbf{a}$ (with one entry for each pair of source and destination states). We assume that the MDP has an infinite horizon and that future rewards are discounted exponentially with a discount factor $\gamma \in [0, 1)$.

In general, the state space $\mathbf{X}$ is more compactly defined in terms of state variables, i.e., $\mathbf{X} = \{X_1, \ldots, X_n\}$. Similarly, the action can be decomposed into action variables $\mathbf{A} = \{A_1, \ldots, A_g\}$.

The optimal value function $V^*$ is defined so that $V^*(\mathbf{x})$ is the maximal value achievable by any action at state $\mathbf{x}$. More precisely, $V^*$ is the solution to (1) below. A stationary policy is a mapping $\pi : \mathbf{X} \mapsto A$, where $\pi(\mathbf{x})$ is the action to be taken at state $\mathbf{x}$. For any value function $V$, we can define the policy obtained by one-step lookahead on $V$: $Greedy[V] = \arg\max_\mathbf{a}[R_\mathbf{a} + P_\mathbf{a}V]$, where the arg max is taken componentwise. The greedy policy for the optimal value function $V^*$ is the optimal policy $\pi^*$.

There are several algorithms for computing the optimal policy (see [17] for a review). One is to solve the Bellman linear program: write $V \in \mathbb{R}^N$ for the value function, so that $V_i$ represents the value of state $i$. Pick any fixed state relevance weights $\alpha \in \mathbb{R}^N$ with $\alpha > 0$; without loss of generality assume $\sum_i \alpha_i = 1$. Then the Bellman LP is

$$\begin{aligned} &\text{Minimize } \alpha \cdot V \\ &V \geq R_\mathbf{a} + \gamma P_\mathbf{a} V \end{aligned} \quad (1)$$

Throughout this paper, inequalities between vectors hold componentwise; e.g., $a \geq b$ means $a_i \geq b_i$ for all $i$. Also, free index variables are by convention universally quantified; e.g., the constraint in (1) holds for all $\mathbf{a}$.

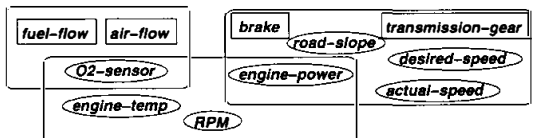

Figure 1: Smart engine control dynamics; e.g., the state variable O2-sensor depends on the action variables fuel-flow and air-flow.

The dual of the Bellman LP is

$$\text{Maximize } \sum_{\mathbf{a}} R_{\mathbf{a}} \cdot \phi_{\mathbf{a}}$$
$$\sum_{\mathbf{a}} \phi_{\mathbf{a}} - \gamma \sum_{\mathbf{a}} P_{\mathbf{a}}^T \phi_{\mathbf{a}} = \alpha, \quad \phi_{\mathbf{a}} \geq 0 \quad (2)$$

The vector $\phi_{\mathbf{a}}$, called the *flow* for action $\mathbf{a}$, can be interpreted as the expected number of times that action $\mathbf{a}$ will be executed in each state (discounted so that future visits count less than present ones). So, the objective of (2) is to maximize total reward for all actions executed, and the constraints say that the number of times we enter each state must be the same as the number of times we leave it. State relevance weights tell us how often we start in each state.

## 4 Hierarchical multiagent factored MDPs

Most interesting MDPs have a very large number of states. For example, suppose that we want to build a smart engine control, whose state at any time is described by the variables shown in Fig. 1. The number of distinct states in such an MDP is exponential in the number of state variables. Similarly, many interesting MDPs have a very large number of actions described by a smaller number of action variables. Even very fast exact algorithms cannot solve such large MDPs in a reasonable amount of time.

Fortunately, in many cases we can group the variables in a large MDP into subsystems that interact in a simple manner. The rounded boxes in Fig. 1 show one possible grouping; we might call the three subsystems fuel-injection, engine-control, and speed-control.

These subsystems overlap with each other on some variables; e.g., fuel-injection and engine-control overlap on O2-sensor. We can consider O2-sensor to be an output of the fuel-injection system and an input to the engine-control.

The advantage of constructing such a grouping is that we can now plan for each subsystem separately. Since there are many fewer variables in each subsystem than there are in the MDP as a whole, we can hope that it will be possible to construct a global plan by stitching together plans for the various subsystems. Of course, if we plan for each subsystem completely separately there's no guarantee that the overall plan will be useful, so we may need to replan for each subsystem several times taking into account the current plans for neighboring subsystems.

In our engine example, we can examine the speed-control subsystem and compute what values of engine-power, brake, and transmission-gear would help us most in keeping actual-speed near desired-speed. While the speed-control system controls transmission-gear and brake directly, it must communicate with the engine-control system to influence the value of engine-power. If the desired value of engine-power turns out to be too expensive to maintain, the speed-control system may have to form a new plan that makes do with the available engine-power.

### 4.1 Subsystem tree MDPs

To formalize the above intuition, we will define a general class of MDPs built from interacting subsystems. In Sec. 5, we will give an algorithm for planning in such MDPs; this algorithm will run efficiently when the number of state variables in the scope of each subsystem is small.

We will start by defining a basic subsystem. On its own, a basic subsystem is just an MDP with factored state and action spaces, but below we will describe how to combine several basic subsystems into a larger MDP by allowing them to share state and action variables. (In particular, an action variable in one subsystem can be a state variable in another; a subsystem's actions are just the variables which can be set by forces external to the subsystem.)

**Definition 4.1 (Basic subsystem)** *A basic subsystem $\mathcal{M}_j$ is an MDP defined by the tuple $(\mathbf{X}_j, \mathbf{A}_j, R_j, P_j)$. The internal state variables $\mathbf{X}_j = \text{Internal}[\mathcal{M}_j]$ and the external state variables $\mathbf{A}_j = \text{External}[\mathcal{M}_j]$ are disjoint sets. The scope of $\mathcal{M}_j$ is $\text{Scope}[\mathcal{M}_j] = \mathbf{X}_j \cup \mathbf{A}_j$. The local reward function $R_j(\mathbf{x}_j, \mathbf{a}_j)$ maps assignments for $\text{Scope}[\mathcal{M}_j]$ to real-valued rewards, and the local transition function $P_j(\mathbf{x}'_j \mid \mathbf{x}_j, \mathbf{a}_j)$ maps assignments for $\text{Scope}[\mathcal{M}_j]$ to a probability distribution over the assignment $\mathbf{x}'_j$ to the internal variables $\mathbf{X}_j$ in the next time step.* ∎

We have divided the scope of a subsystem into internal variables and external variables. Each subsystem knows the dynamics of its internal variables, and can therefore reason about value functions that depend on these variables. External variables are those which the subsystem doesn't know how to influence directly; a subsystem may form opinions about which values of the external variables would be helpful or harmful, but cannot perform Bellman backups for value functions that depend on external variables.

We will write $R_{j,\mathbf{a}_j}$ for the vector of rewards associated with a setting $\mathbf{a}_j$ for $\mathbf{A}_j$. $R_{j,\mathbf{a}_j}$ has one component for each setting of $\mathbf{X}_j$. Similarly, we will write $P_{j,\mathbf{a}_j}$ for the transition matrix associated with setting $\mathbf{A}_j$ to $\mathbf{a}_j$; each column of $P_{j,\mathbf{a}_j}$ corresponds to a single setting of $\mathbf{X}_j$ at time $t$, while each row corresponds to a setting of $\mathbf{X}_j$ at time $t+1$.

Given several basic subsystems, we can collect them together into a subsystem tree. Right now we are not imposing any consistency conditions on the subsystems, but we will do so in a little while.

**Definition 4.2 (Subsystem tree)** *A subsystem tree $\mathcal{M}$ is a set of local subsystem MDPs $\mathcal{M}_1, \ldots, \mathcal{M}_m$ together with a tree parent function $\mathcal{M}_i = \text{Parent}[\mathcal{M}_j]$. Without loss of generality take $\mathcal{M}_1$ to be the root of the tree. We will write $\text{Parent}[\mathcal{M}_1] = \emptyset$, and we will define $\text{Children}[\mathcal{M}_j]$ in the usual way. The internal variables of the tree, $\text{Internal}[\mathcal{M}]$, are those variables which are internal to any $\mathcal{M}_j$; the external variables, $\text{External}[\mathcal{M}]$, are the variables which are in the scope of some subsystem but not internal to any subsystem. Finally, the common (separator) variables between* *a subsystem and its parent are denoted by* $\text{SepSet}[\mathcal{M}_j] = \mathcal{S}_j = \text{Scope}[\mathcal{M}_j] \cap \text{Scope}[\text{Parent}[\mathcal{M}_j]]$. ∎

There are two consistency conditions we need to enforce on a subsystem tree to make sure that it defines an MDP. The first says that every subsystem that references a variable $X_i$ must be connected to every other subsystem referencing $X_i$, and the second says that neighboring subsystems have to agree on transition probabilities.

**Definition 4.3 (Running intersection property)** *A subsystem tree* $\mathcal{M}$ *satisfies the* running intersection property *if, whenever a variable* $X_i$ *is in both* $\text{Scope}[\mathcal{M}_j]$ *and* $\text{Scope}[\mathcal{M}_k]$, *then* $X_i \in \text{Scope}[\mathcal{M}_l]$ *for every subsystem* $\mathcal{M}_l$ *in the path between* $\mathcal{M}_j$ *and* $\mathcal{M}_k$ *in* $\mathcal{M}$. *Similarly, internal variables must satisfy the same condition.* ∎

**Definition 4.4 (Consistent dynamics)** *A subsystem tree* $\mathcal{M}$ *has* consistent dynamics *if pairs of subsystems agree on the dynamics of their joint variables. Specifically, if* $\mathbf{z}$ *is an assignment to* $\text{Scope}[\mathcal{M}_j] \cup \text{Scope}[\mathcal{M}_k]$, *and if* $\mathbf{x}'$ *is an assignment to* $\text{Internal}[\mathcal{M}_j] \cap \text{Internal}[\mathcal{M}_k]$ *in the next time step, then*

$$\sum_{[\mathbf{x}'_k|\mathbf{x}']} P_k(\mathbf{x}'_k \mid \mathbf{z}) = \sum_{[\mathbf{x}'_j|\mathbf{x}']} P_j(\mathbf{x}'_j \mid \mathbf{z})$$

*Here the sum* $[\mathbf{x}'_k \mid \mathbf{x}']$ *runs over all assignments* $\mathbf{x}'_k$ *to* $\text{Internal}[\mathcal{M}_k]$ *consistent with* $\mathbf{x}'$. ∎

A subsystem tree which satisfies the running intersection property and has consistent dynamics is called a *consistent subsystem tree*. For the rest of this paper, all subsystem trees will be assumed to be consistent. Given a consistent subsystem tree, we can construct a plain old MDP, which defines the dynamics associated with this tree:

**Lemma 4.5 (Equivalent MDP)** *From a consistent subsystem tree* $\mathcal{M}$ *we can construct a well-defined equivalent MDP,* $\text{MDP}[\mathcal{M}] = (\mathbf{X}, \mathbf{A}, R, P)$. *The state variable set is* $\mathbf{X} = \text{Internal}[\mathcal{M}]$. *The action variable set is* $\mathbf{A} = \text{External}[\mathcal{M}]$. *The reward function is the sum of all subsystem rewards* $R = \sum_{j=1}^{m} R_j$. *The transition probabilities for a given state and action* $\mathbf{z} = (\mathbf{x}, \mathbf{a})$ *are*

$$P(\mathbf{x} \mid \mathbf{z}) = \frac{\prod_{j=1}^{m} P_j(\mathbf{x}_j \mid \mathbf{z}_j)}{\prod_{k=2}^{m} \sum_{[\mathbf{x}_k|\mathbf{s}_k]} P_k(\mathbf{x}_k \mid \mathbf{z}_k)}$$

*where* $\mathbf{z}_j$ *is the restriction of* $\mathbf{z}$ *to the scope of* $\mathcal{M}_j$; $\mathbf{x}_j$ *is the restriction of* $\mathbf{x}$ *to the internal variables of* $\mathcal{M}_j$; $\mathbf{s}_k$ *is the restriction of* $\mathbf{z}$ *to variables in* $\text{SepSet}[\mathcal{M}_k]$; *and* $[\mathbf{x}_k \mid \mathbf{s}_k]$ *denotes the set of assignments to* $\mathbf{x}_k$ *consistent with* $\mathbf{s}_k$. ∎

In our engine example, the rounded boxes in Fig. 1 are the scopes of the three subsystems. The O2-sensor variable is an external variable for the engine-control system, and engine-power is an external variable for the speed-control system; all other variables of each subsystem are internal. (Note: the way we have described this example, each variable is internal to only one subsystem; this is not necessary so long as we maintain the running intersection property.)

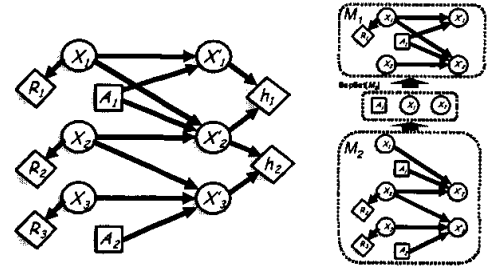

Figure 2: Multiagent factored MDP with basis functions $h_1$ and $h_2$ (left) represented as a hierarchical subsystem tree (right).

### 4.2 Hierarchical subsystem trees

In our engine example, we might decide to add several more variables describing the internal state of the engine. In this case the engine-control subsystem would become noticeably more complicated than the other two subsystems in our decomposition. Since our decomposition algorithm below allows us to use any method we want for solving the subsystem MDPs, we could use another subsystem decomposition to split engine-control into sub-subsystems. For example, we might decide that we should break it down into a distributor subsystem, a power-train subsystem, and four cylinder subsystems. This recursive decomposition would result in a tree of subsystem trees. We call such a tree of trees a *hierarchical subsystem tree*.

Hierarchical subsystem trees are important for two reasons. First, they are important for knowledge representation: it is easier to represent and reason about hierarchies of subsystems than flat collections. Second, they are important for reuse: if the same subsystem appears several times in our model, we can reuse plans from one copy to speed up planning for other copies (see Sec. 9 for details).

### 4.3 Relationship to factored MDPs

*Factored MDPs* [2] allow the representation of large structured MDPs by using a *dynamic Bayesian network* [5] as the transition model. Guestrin et al. [8] used factored MDPs for multiagent planning. They presented a planning algorithm which approximates the value function of a factored MDP with *factored linear value functions* [10]. These value functions are a weighted linear combination of basis functions where each basis function is restricted to depend only on a small subset of state variables.

The relationship between factored MDPs and the hierarchical decomposition described in this paper is analogous to the one between standard Bayesian networks and Object-Oriented Bayesian networks [11]. In terms of representational power, hierarchical multiagent factored MDPs are equivalent to factored MDPs with factored linear value functions. That is, a factored MDP associated with some choice of basis functions can be easily transformed into a subsystem tree with a particular choice of subsystems and vice-versa.[1] This transformation involves the backprojection of the basis functions [10] and the triangulation of the resulting dependency graph into a clique tree [13]. For example, consider the factored MDP on the left of Fig. 2, with basis functions $\{h_1, h_2\}$ represented as diamonds in the next time step. The figure on the right represents an equivalent subsystem tree, where each local dynamical system is represented by a small part of the global DBN.

---
[1] For some basis function choices, the transformation from factored MDPs to subsystem trees may also imply an approximate solution of the basic subsystem MDPs.

However, the hierarchical model offers some advantages: First, we can specify a simple, completely *distributed* planning algorithm for this model (Sec. 5). Second, the hierarchical model allows us to reuse plans generated for two equivalent subsystems. Third, the knowledge engineering task is simplified as systems can be built from a library of subsystems. Finally, even in collaborative multiagent settings, each agent may not want to reveal private information; e.g., in a factory, upper management may not want to reveal the salary of one section manager to another. Using our hierarchical approach, each subsystem could be associated with an agent, and each agent would only have access to its local model and reward function.

## 5  Solving hierarchical factored MDPs

We need to address some problems before we can solve the Bellman LP for an MDP represented as a subsystem tree: the LP has exponentially many variables and constraints, and it has no separation between subsystems. This section describes our solutions to these problems.

### 5.1  Approximating the Bellman LP

Consider the MDP obtained from a subsystem tree $\mathcal{M}$ according to Lemma 4.5. This MDP's state and action spaces are exponentially large, with one state for each assignment $\mathbf{x}$ to $\{X_1, \ldots, X_n\}$ and one action for each assignment $\mathbf{a}$ to $\{A_1, \ldots, A_g\}$; so, optimal planning is intractable. We use the common approach of restricting attention to value functions that are compactly represented as a linear combination of basis functions $\{h_1, \ldots, h_k\}$. We will write $\mathbf{w} \in \mathbb{R}^k$ for the weights of our linear combination and $H$ for the matrix whose columns are the basis functions; so, our representation of the value function is $V = H\mathbf{w}$.

As proposed by Schweitzer and Seidmann [19], we can substitute our approximate value function representation into the Bellman LP (1):

$$\begin{aligned} \text{Minimize } & \alpha \cdot H\mathbf{w} \\ & H\mathbf{w} \geq R_\mathbf{a} + \gamma P_\mathbf{a} H\mathbf{w} \end{aligned} \quad (3)$$

There is, in general, no guarantee on the quality of the approximation $V = H\mathbf{w}$, but recent work of de Farias and Van Roy [4] provides some analysis of the error relative to that of the best possible approximation in the subspace, and some guidance as to selecting the state relevance weights $\alpha$ so as to improve the quality of the approximation.

We will choose the basis $H$ to reflect the structure of $\mathcal{M}$: we will allow ourselves complete flexibility to represent the value function $V_j$ of each subsystem $\mathcal{M}_j$, but we will approximate the global value function $V$ by the sum of the subsystem value functions.[2] If $\mathcal{M}_j$ is itself a subsystem tree, we will further approximate the global value function by decomposing $V_j$ recursively into a sum of its sub-subsystem value functions; but for simplicity of notation, we will assume that $\mathcal{M}$ has been flattened so that the $\mathcal{M}_j$s are all basic subsystem MDPs.

More specifically, let $H_j$ be the basis for $\mathcal{M}_j$ within $\mathcal{M}$. In other words, let the $i$th column of $H_j$ be an indicator function over assignments to Internal[$\mathcal{M}$] which is 1 when Internal[$\mathcal{M}_j$] is set to its $i$th possible setting, and 0 otherwise. Then we can write $V = \sum_j H_j V_j$. Substituting this approximation into (3) yields

$$\begin{aligned} \text{Minimize } & \sum_j \alpha \cdot H_j V_j \\ & \sum_j H_j V_j \geq R_\mathbf{a} + \sum_j \gamma P_\mathbf{a} H_j V_j \end{aligned} \quad (4)$$

### 5.2  Factoring the approximate LP

The substitution (4) dramatically reduces the number of free variables in our linear program: instead of one variable for each assignment to Internal[$\mathcal{M}$], we now have one variable for each assignment to Internal[$\mathcal{M}_j$] for each $j$. The number of constraints, however, remains the same: one for each assignment to Scope[$\mathcal{M}$]. Fortunately, Guestrin et al. [8] proposed an algorithm that reduces the number of constraints in a factored MDP by a method analogous to variable elimination.

Their algorithm introduces an extra variable $S_j(\mathbf{z}_j)$ (called a message variable) for each possible assignment $\mathbf{z}_j$ to each separating set SepSet[$\mathcal{M}_j$]. To simplify notation we will sometimes write extra arguments to $S_j$; for instance, if $\mathbf{x}_j, \mathbf{a}_j$ is an assignment to Scope[$\mathcal{M}_j$], it determines the value of $\mathbf{z}_j$ and so we can write $S_j(\mathbf{x}_j, \mathbf{a}_j)$. We will also introduce dummy variables $U_{j, \mathbf{a}_j}$ to represent the total influence of all of the $S$'s on subsystem $\mathcal{M}_j$ under action $\mathbf{a}_j$. $U_{j, \mathbf{a}_j}$ is a vector with one component $U_{j, \mathbf{x}_j, \mathbf{a}_j}$ for each assignment $\mathbf{x}_j$ to Internal[$\mathcal{M}_j$].

With this notation, the Guestrin et al. algorithm reduces (4) to the following LP:

$$\begin{aligned} \text{Minimize } & \sum_j \alpha \cdot H_j V_j \\ & V_j \geq R_{j, \mathbf{a}_j} + U_{j, \mathbf{a}_j} + \gamma P_{j, \mathbf{a}_j} V_j \\ & U_{j, \mathbf{x}_j, \mathbf{a}_j} = \sum_{k \in \text{ch}_j} S_k(\mathbf{x}_j, \mathbf{a}_j) - S_j(\mathbf{x}_j, \mathbf{a}_j) \end{aligned} \quad (5)$$

(The set ch$_j$ contains all $k$ such that $\mathcal{M}_k \in$ *Children*[$\mathcal{M}_j$].) This LP has many fewer constraints than (4): there is one inequality for each $j, \mathbf{x}_j, \mathbf{a}_j$ instead of one for each $\mathbf{x}, \mathbf{a}$.

If we are willing to assume a centralized planner with knowledge of the details of every subsystem, we are now done: we can just construct (5) and hand it to a linear program solving package.[3] In order to achieve distributed planning, though, we need to break (5) into pieces which refer only to a single subsystem and its neighbors. (In fact, even if we are relying on a central planner, we may still want to break (5) into subsystems to give ourselves the flexibility to solve each subsystem with a different algorithm.)

---

[2] We are not restricting $V_j$ to be the value function which would be optimal if $\mathcal{M}_j$ were isolated.

[3] Actually, we need one more assumption: the state relevance weights $\alpha$ must factor along subsystem lines so that we can compute $H_j^T \alpha$ efficiently. Equivalently, we can pick subsystem weight vectors $\bar{\alpha}_j$ that satisfy a condition analogous to consistent dynamics, then define $\alpha$ so that $\bar{\alpha}_j = H_j^T \alpha$.

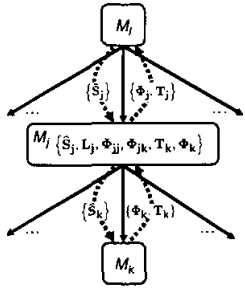

Figure 3: Message passing algorithm from the point of view of subsystem $\mathcal{M}_j$: it maintains a set of local flows $\Phi_{jj}$ and $\Phi_{jk}$ with corresponding local values $L_j$. $\mathcal{M}_j$ receives the current reward message $\widehat{S}_j$ from its parent $\mathcal{M}_l$. Additionally, each child subsystem $\mathcal{M}_k$ sends a message to $\mathcal{M}_j$ containing the flows $\Phi_k$ over their separator variables, along with the corresponding total values $T_k$ of the subtree rooted at $\mathcal{M}_k$. $\mathcal{M}_j$ combines this information to compute the new reward messages $\widehat{S}_k$ for each child, along with the flow message $\Phi_j$ and total value $T_j$ sent to $\mathcal{M}_l$.

### 5.3 Reward sharing

If we fix all message variables $S_k(\mathbf{z}_k)$, the LP in (5) splits into many small pieces. An individual piece looks like:

$$\text{Minimize } \bar{\alpha}_j \cdot V_j \qquad (6)$$
$$V_j \geq R_{j,\mathbf{a}_j} + U_{j,\mathbf{a}_j} + \gamma P_{j,\mathbf{a}_j} V_j$$

Here we have written $\bar{\alpha}_j$ for $H_j^T \alpha$. The vectors $U_{j,\mathbf{a}_j}$ are now constants.

By comparing (6) to the standard Bellman LP (1), we can see that it represents an MDP with rewards $R_{j,\mathbf{a}_j} + U_{j,\mathbf{a}_j}$ and transition probabilities $P_{j,\mathbf{a}_j}$. We will call this MDP the *stand-alone MDP* for subsystem $\mathcal{M}_j$.

Viewing (6) as an MDP allows us to interpret $U_{j,\mathbf{a}_j}$ as a vector of adjustments to the reward for subsystem $\mathcal{M}_j$. These adjustments depend on the separating sets between $\mathcal{M}_j$ and its neighbors in the subsystem tree.

By looking at the definition of $U_{j,\mathbf{a}_j}$ in (5), we can see how a good setting for the message variables encourages cooperative behavior. The message variable $S_j(\mathbf{z}_j)$ reduces the reward of $\mathcal{M}_j$ in states where $\mathbf{z}_j$ is true, but increases the reward of $\mathcal{M}_j$'s parent by the same amount. So, if $\mathcal{M}_j$ benefits from being in states where $\mathbf{z}_j$ is true, we can increase $S_j(\mathbf{z}_j)$ to cause $\mathcal{M}_j$'s parent to seek out those states. Conversely, if the $\mathbf{z}_j$ states hurt $\mathcal{M}_j$'s parent, we can decrease $S_j(\mathbf{z}_j)$ to encourage $\mathcal{M}_j$ to avoid them.

This interpretation is an important feature of our algorithm. We have taken a complex planning problem which may have many strong couplings between subsystems, and defined a small number of message variables which allow us to reduce the global coordination problem to the problem of finding an appropriate reward-sharing plan.

Our algorithm is also linked to reward shaping [18]. In reinforcement learning, it is common to add fictitious shaping rewards to the system to speed up learning. The purpose of the reward message in our approach is to encourage coordination rather than fast learning. Nonetheless, the reward messages do shape the subsystems' policies to conform to a globally consistent strategy.

### 5.4 Algorithm description

In this subsection we will describe our algorithm for finding a good reward-sharing plan. The algorithm is guaranteed to converge in a finite number of iterations; at termination we will have found an exact solution to (5).

Our algorithm maintains several sets of variables at each subsystem in the tree. These variables represent messages passed between neighboring subsystems in the tree. All messages are about one of two topics: rewards or expected frequencies (flows). Flow messages pass up the tree from child to parent, generating reward messages in their wake. These reward messages cause neighboring subsystems to replan, which in turn generates more flow messages. Fig. 3 illustrates the messages exchanged between subsystems.

Reward messages allow our algorithm to encourage cooperative behavior by rewarding actions which help neighboring subsystems and punishing actions which hurt them. Flow messages tell subsystems about the policies their neighbors are considering; they let subsystems know what assignments to $S_k$ their neighbors have figured out how to achieve (and at what cost).

The first set of message variables is $\widehat{S}_j(\mathbf{z}_j)$, the most recent reward message received by $\mathcal{M}_j$ from its parent. The second set consists of $\widehat{S}_k(\mathbf{z}_k)$, the most recent reward-sharing plan sent to $\mathcal{M}_j$'s children. The remaining sets of variables keep track of important statistics about the policies found so far. The algorithm uses these statistics to help generate new policies for various parts of the subsystem tree.

We keep track of policies both for individual subsys-tems and for groups of subsystems. For each individual subsystem $\mathcal{M}_j$, we keep a vector $L_j$ whose $i$th component is the local expected discounted reward of the $i$th policy we computed for $\mathcal{M}_j$. We also keep matrices $\Phi_{jk}$ for every separating set $\mathcal{S}_k$ that touches $\mathcal{M}_j$. The $i$th row of $\Phi_{jk}$ tells us how often the $i$th policy sets the variables of $\mathcal{S}_k$ to each of their possible values.

---

1. Initialization:
   — $t \leftarrow 1$
   — For all subsystems $\mathcal{M}_j$, $\Phi_j \leftarrow \emptyset$, $T_j \leftarrow \emptyset$, $L_j \leftarrow \emptyset$, and $\widehat{S}_j \leftarrow 0$.
   — For all $\mathcal{M}_j$ and separating sets $\mathcal{S}_k$ touching $\mathcal{M}_j$, $\Phi_{jk} \leftarrow \emptyset$

2. For each subsystem $\mathcal{M}_j$, if $\Phi_{jk}$ or $\Phi_k$ for $k \in \text{ch}_j$ changed in the last iteration:
   — Solve the reward message LP (11) to find new values for the message variables $\widehat{S}_k$ of each separating set between $\mathcal{M}_j$ and its children.
   — If the reward message LP was bounded, use its value and dual variables to add a new entry to $T_j$ according to (9). Also add a new row to $\Phi_j$ according to (10).

3. For every $j$, if $\mathcal{M}_j$ depends on a reward message which changed in step 2, solve its stand-alone MDP (6). Add a new entry to $L_j$ according to (7). For every separating set $\mathcal{S}_k$ which touches $\mathcal{M}_j$, add a new row to $\Phi_{jk}$ as in Eq. (8).

4. If an $\widehat{S}$ or a $\Phi$ has changed, set $t \leftarrow t + 1$ and go to 2.

Figure 4: Planning algorithm for subsystem trees.

If $\phi$ is a feasible flow for $\mathcal{M}_j$, composed of one vector $\phi_\mathbf{a}$ for each action $\mathbf{a}$, then its component in $L_j$ is

$$\sum_\mathbf{a} R_\mathbf{a} \cdot \phi_\mathbf{a} \qquad (7)$$

This is the reward for $\phi$ excluding contributions from message variables. The corresponding row of $\Phi_{jk}$ has one element for every assignment $\mathbf{z}_k$ to the variables of $\mathcal{S}_k$. This element is:

$$\sum_{[\mathbf{x},\mathbf{a}|\mathbf{z}_k]} \phi_\mathbf{a}(\mathbf{x}) \qquad (8)$$

This is the marginalization of $\phi$ to the variables of $\mathcal{S}_k$.

Consider now the subtree rooted at $\mathcal{M}_j$. For this subtree we keep a vector of subtree rewards $T_j$ and a matrix of frequencies $\Phi_j$. A new policy for a subtree is a mixture of policies for its children and its root; we can represent such a mixed policy with some vectors of mixture weights. We will need one vector of mixture weights for each child (call them $p_k$ for $k \in \text{ch}_j$) and one for $\mathcal{M}_j$ (call it $p_j$). Each element of $p_j$ and the $p_k$s is a weight for one of the previous policies we have computed.

Given a set of mixture weights, compute a new entry for $T_j$ in terms of $\mathcal{M}_j$ and the trees rooted at $\mathcal{M}_j$'s children:

$$\sum_k p_k \cdot T_k + p_j \cdot L_j \qquad (9)$$

This is the expected discounted reward for our mixed policy, excluding contributions from message variables. We can also compute the corresponding new row for $\Phi_j$: it is

$$p_j^T \Phi_{jj} \qquad (10)$$

This is how often our mixed policy sets the variables of $\mathcal{S}_j$ to each of their possible values.

Our algorithm alternates between two ways of generating new policies. The simpler way is to solve a stand-alone MDP for one of the subsystems; this will generate a new policy if we have changed any of the related reward message variables. The second way is to solve a reward message linear program; this LP updates some reward message variables and also produces a set of mixture weights for use in equations (9) and (10).

There is a reward message LP for the subtree rooted at any non-leaf subsystem $\mathcal{M}_j$. Let the index $k$ run over $\text{ch}_j$; then we can write the LP as:

$$\begin{aligned}
&\text{Minimize } \theta_j + \sum_k \theta_k \\
&\mathbf{1}\theta_j \geq L_j - \Phi_{jj}\widehat{S}_j + \sum_k \Phi_{jk} S_k \\
&\mathbf{1}\theta_k \geq T_k - \Phi_k S_k
\end{aligned} \qquad (11)$$

The solution of this LP tells us the value of the new reward messages $\widehat{S}_k$ to be sent to $\mathcal{M}_j$'s children. To obtain mixture weights, we can look at the dual of (11):

$$\begin{aligned}
&\text{Maximize } \sum_k p_k \cdot T_k + p_j \cdot (L_j - \Phi_{jj}\widehat{S}_j) \\
&\Phi_k^T p_k = \Phi_{jk}^T p_j \\
&\mathbf{1}^T p_j = 1 \quad \mathbf{1}^T p_k = 1 \\
&p_j \geq 0 \quad p_k \geq 0
\end{aligned} \qquad (12)$$

These mixture weights are used to generate the message to be sent to $\mathcal{M}_j$'s parent. Fig. 4 brings all of these ideas together into a message-passing algorithm which propagates information through our subsystem tree. The following theorem guarantees the correctness of our algorithm; for a proof see Sec. 7.

**Theorem 5.1 (Convergence and correctness)** *Let $\mathcal{M}$ be a subsystem tree. The distributed algorithm in Fig. 4 converges in a finite number of iterations to a solution of the global linear program (4) for $\mathcal{M}$.* ∎

While Fig. 4 describes a specific order in which agents update their policies, other orders will work as well. In fact, Thm. 5.1 still holds so long as every agent eventually responds to every message sent to it.

## 6 An example

Before we justify our algorithm, we will work through a simple example. This example demonstrates how to determine which messages agents need to pass to each other, as well as how to interpret those messages.

### 6.1 A simple MDP

Our example MDP has 2 binary state variables, $x$ and $y$, and 2 binary action variables, $a$ and $b$. The state evolves according to $x_{t+1} = a_t$ and $y_{t+1} = b_t \wedge x_t$. Our per-step reward is $10y - 3x$ and our discount factor is $\gamma = 0.9$. That means there is a tension between wanting $x = 0$ to avoid an immediate penalty and wanting $x = 1$ to allow $y = 1$ later. The exact value function for this MDP is $(54, 64, 60, 70)$ for the states $xy = (00, 01, 10, 11)$.

We will decompose our MDP into 2 subsystems, one with internal variable $x$ and external variable $a$, and one with internal variable $y$ and external variables $x$ and $b$. This decomposition cannot represent all possible value functions: it is restricted to the family $V(x, y) = V_1(x) + V_2(y)$, which has only three independent parameters. However, the exact value function is a member of this family (with $V_1 = (54, 60)$ and $V_2 = (0, 10)$), so our decomposition algorithm will find it.

### 6.2 The LP and its decomposition

With the above definitions, we can write out the full linear program for computing $V(x, y)$:

$$\begin{aligned}
&\text{Minimize } \sum_x V_1(x) + \sum_y V_2(y) \\
&V_1(x) + V_2(y) \geq 10y - 3x + \gamma(V_1(a) + V_2(x \wedge b))
\end{aligned}$$

There are 16 constraints in the full LP (4 states × 4 actions). We can reduce that to 10 by introducing two new variables

$$\begin{array}{cc|cc|cc}
1 & 0 & 1-\gamma & 0 & 0 & 0 \\
1 & 0 & 1 & -\gamma & 0 & 0 \\
0 & 1 & -\gamma & 1 & 0 & 0 \\
0 & 1 & 0 & 1-\gamma & 0 & 0 \\
\hline
-1 & 0 & 0 & 0 & 1-\gamma & 0 \\
-1 & 0 & 0 & 0 & -\gamma & 1 \\
0 & -1 & 0 & 0 & 1-\gamma & 0 \\
0 & -1 & 0 & 0 & 1 & -\gamma \\
0 & -1 & 0 & 0 & -\gamma & 1 \\
0 & -1 & 0 & 0 & 0 & 1-\gamma
\end{array}$$

Figure 5: Constraint matrix for the example MDP after variable elimination. Columns correspond to the variables $S(0), S(1), V_1(0), V_1(1), V_2(0), V_2(1)$ in that order.

$S(x)$. $S(1)$ represents $\min_{y,b}[V_2(y) - \gamma V_2(b) - 10y]$ (the minimum of the part of the constraint depending on $y$ and $b$, if $x$ is 1) and similarly for $S(0)$. Now we write our LP as

$$\text{Minimize } \sum_x V_1(x) + \sum_y V_2(y)$$
$$V_1(x) \geq \gamma V_1(a) - 3x - S(x)$$
$$S(0) \leq V_2(y) - \gamma V_2(0) - 10y$$
$$S(1) \leq V_2(y) - \gamma V_2(b) - 10y$$

This LP has the constraint matrix shown in Fig. 5. As indicated, the matrix has a simple block structure: two variables $S(x)$ appear in all constraints; two variables $V_1(x)$ appear only in the subproblem above the horizontal line; and two variables $V_2(y)$ appear only in the subproblem below the horizontal line. This block structure is what allows our algorithm to plan separately for the two subproblems.

### 6.3 Execution of the decomposition algorithm

The decomposition algorithm starts out with $S(x) = 0$ (no reward sharing). So, $\mathcal{M}_2$ always picks $x = 1$, since that allows it to set $y = 1$ and capture a reward of 10 on each step. Similarly, $\mathcal{M}_1$ sees no benefit to visiting its $x = 1$ state, so it heads for $x = 0$ to avoid a reward of -3 per step.

Each of these two policies results in a new constraint for our message LP. For example, $\mathcal{M}_2$'s policy tells us that $\theta_2 \geq 95 + \frac{1}{1-\gamma}S(1)$, since it achieves an average reward of 95 when $S(1) = 0$ and always sets $x = 1$.

Adding the new constraints and re-solving the message LP tells us that $\mathcal{M}_1$ and $\mathcal{M}_2$ disagree about how often $x$ should equal 1, and suggests putting a premium on $x = 1$ for $\mathcal{M}_1$ and the corresponding penalty on $x = 1$ for $\mathcal{M}_2$. (The message LP is unbounded at this point, so the size of the premium/penalty is arbitrary so long as it is large.) As we would expect, the two MDPs react to this change in rewards by changing their policies: in step 2 $\mathcal{M}_1$ decides it will set $x = 1$ as often as possible, and $\mathcal{M}_2$ decides it will set $x = 0$ as often as possible.

With the constraints derived from these new policies, the message LP decides that it will give $\mathcal{M}_1$ a reward of 9 for setting $x = 1$, and charge $\mathcal{M}_2$ the corresponding penalty. With this new reward structure, the two MDPs can now compute what will turn out to be their final policies: $\mathcal{M}_1$ will set $x = 1$ as often as possible despite the one-step penalty of -3, thereby allowing $\mathcal{M}_2$ to set $y = 1$ and achieve a reward of 10 on each step. Summing the resulting value functions for the two subproblems gives the true globally-optimal value function for the overall MDP, and further steps of our algorithm do not change this result.

## 7 Algorithm justification

We can derive the algorithm in Fig. 4 by performing a sequence of nested Benders decompositions on the linear program (5). This section reviews Benders decomposition, then outlines how to apply it to (5) to produce our algorithm. Since Benders decomposition is correct and finitely convergent, this section is a proof sketch for Thm. 5.1.

### 7.1 Benders decomposition

Benders decomposition [1] solves LPs of the form

$$\begin{aligned}\text{Minimize } & a \cdot x + b \cdot y \\ & Cx + Dy \geq k \end{aligned} \quad (13)$$

by repeatedly solving subproblems where the value of $x$ is fixed. It is useful mainly when the subproblems are easier to solve than the problem as a whole, perhaps because the matrix $D$ is block diagonal or has some other special structure. (Other types of constraints besides $\geq$ are also possible.) We will call $x$ the master variable and $y$ the slave variable. If we fix $x = \hat{x}$, we get the subproblem:

$$\begin{aligned}\text{Minimize } & b \cdot y \\ & Dy \geq k - C\hat{x}\end{aligned} \quad (14)$$

Writing $\theta(\hat{x})$ for the optimal value of this subproblem, we reduce (13) to: Minimize $a \cdot x + \theta(x)$. The dual of (14) is

$$\begin{aligned}\text{Maximize } & \phi \cdot (k - C\hat{x}) \\ & D'\phi = b, \quad \phi \geq 0\end{aligned} \quad (15)$$

Note that the feasible region of (15) is independent of $\hat{x}$. If we have a feasible solution $\hat{\phi}$ to (15), it provides a lower bound on the subproblem value by plugging $\hat{\phi}$ into the objective of (15): $\theta(x) \geq \hat{\phi} \cdot (k - Cx)$. If we have several feasible solutions $\phi_1, \phi_2, \ldots$, each one provides a lower bound on $\theta(x)$. So, we can approximate the reduced version of (13) with

$$\begin{aligned}\text{Minimize } & a \cdot x + \theta \\ & \theta \geq \phi_i \cdot (k - Cx)\end{aligned} \quad (16)$$

The Benders algorithm repeatedly solves (16) to get a new value of $x$, then plugs that value of $x$ into (15) and solves for a new $\phi_i$. The process is guaranteed to converge in finitely many iterations.

### 7.2 Decomposing the factored LP

We can pick any set of message variables to start our Benders decomposition. Suppose we pick $S_k(\mathbf{z}_k)$ for all $\mathcal{M}_k$s which are children of the root $\mathcal{M}_1$. These $S_k$s will be master variables, and all remaining LP variables will be slaves.

Fixing these message variables to $\widehat{S}_k$ separates the root from its children. So, the Benders subproblem will split into several separate pieces which we can solve independently. One piece will be just the stand-alone MDP for the root subsystem, and each other piece will contain a whole subsystem tree rooted at one of $\mathcal{M}_1$'s children.

Using this decomposition, our master becomes:

$$\text{Minimize } \theta_j + \sum_{k \in \text{ch}_j} \theta_k$$
$$\text{Constraints in } \mathcal{Q}_j \quad (17)$$

where $\theta_j$ is the objective of the root stand-alone MDP and the $\theta_k$s are the objectives of the LPs for the subtrees rooted at each child $\mathcal{M}_k$. The set $\mathcal{Q}_j$ contains the constraints received from each subproblem.

First, consider the stand-alone MDP for the root. The dual of its Bellman LP (6) is:

$$\text{Maximize } \sum_{\mathbf{a}_j} \left( R_{j,\mathbf{a}_j} + \widehat{U}_{j,\mathbf{a}_j} \right) \cdot \phi_{\mathbf{a}_j}$$
$$\sum_{\mathbf{a}} \phi_{\mathbf{a}_j} - \gamma \sum_{\mathbf{a}} P_{j,\mathbf{a}_j}^T \phi_{\mathbf{a}_j} = \bar{\alpha}_j \quad \phi_{\mathbf{a}_j} \geq 0$$

where $\widehat{U}_{j,\mathbf{a}_j}$ is a constant vector specified by the choice of $\widehat{S}_k$. We note that the $\widehat{S}_k$s appear only in the objective. Thus, any policy for this subsystem will yield flows $\widehat{\phi}_{\mathbf{a}_j}$ which are feasible for any setting of the $\widehat{S}_k$s. These flows will, in turn, yield a constraint for the LP in (17):

$$\theta_j \geq \sum_{\mathbf{a}_j} R_{j,\mathbf{a}_j} \cdot \widehat{\phi}_{\mathbf{a}_j} + \sum_{\mathbf{a}_j} U_{j,\mathbf{a}_j} \cdot \widehat{\phi}_{\mathbf{a}_j} \quad (18)$$

The first part of the constraint is the value of the policy associated with $\widehat{\phi}_{\mathbf{a}_j}$, which we stored as an entry in $L_j$ in Sec. 5.4. The second part is the product of the flows (which we stored as a row of $\Phi_{jk}$) with the reward messages.

Now, let's turn our attention to the LP for the subtree rooted at a child $\mathcal{M}_k$. By taking the dual of this LP, we will obtain a constraint of the same form as the one in Eq. (18). However, the two terms will correspond to the value of the whole subtree and the flows of the whole subtree, respectively. Fortunately, we can compute these terms with another Benders decomposition that separates $\mathcal{M}_k$ from its children. This recursive decomposition gives us the quantities we called $T_k$ and $\Phi_k$. Note that we do not need the complete set of flows, but only the marginals over the variables in $\mathcal{M}_j$; so, we can compute $\Phi_k$ locally by Eq. (10). The proof is concluded by induction.

## 8 Hierarchical action selection

Once the planning algorithm has converged to a solution $V_j$ for each subsystem $\mathcal{M}_j$, we need a method for selecting the greedy action associated with the global value function $V = \sum_j H_j V_j$. We might try to compute the best action by enumerating all actions and comparing them. Unfortunately, our action space is exponentially large, making this approach infeasible. However, we can exploit the subsystem tree structure to select an action efficiently [8].

Recall that we are interested in finding the greedy action which maximizes the $Q$ function. Our value function is decomposed as the sum of local value functions over subsystems. This decomposition also implies a decomposition of the $Q$ function: $Q = \sum_j Q_j$, where: $Q_j(\mathbf{x}_j, \mathbf{a}_j) = R_j(\mathbf{x}_j, \mathbf{a}_j) + \gamma \sum_{\mathbf{x}'_j} P_j(\mathbf{x}'_j \mid \mathbf{x}_j, \mathbf{a}_j) V_j(\mathbf{x}'_j)$.

Note that some of the external variables $\mathbf{a}_j$ will be internal to some other subsystem, while others correspond to actual action choices. More specifically, for each subsystem $\mathcal{M}_j$, divide its variables into those which are internal to some subsystem in $\mathcal{M}$ (state variables) and those which are external to all subsystems (action variables). Write $\mathbf{y}_j$ for the former and $\mathbf{b}_j$ for the latter.

At each time step $t$, $\mathcal{M}_j$ observes the current value of $\mathbf{y}_j$. (All of these variables are either internal or external to $\mathcal{M}_j$, so a subsystem never needs to observe variables outside its scope.) Subsystem $\mathcal{M}_j$ then instantiates the state-variables part of $Q_j$ to $\mathbf{y}_j$, generating a new local $Q_j$ function, denoted by $Q_j^{(t)}(\mathbf{b}_j)$, which only depends on local action variables $\mathbf{b}_j$.

The subsystems must now combine their local $Q_j$ functions to decide which action is globally greedy, i.e., which action $\mathbf{b}$ maximizes $\sum_j Q_j^{(t)}([\mathbf{b}_j \mid \mathbf{b}])$. They can do so by making two passes over the subsystem tree, one upwards and one downwards. If the parent of $\mathcal{M}_j$ is $\mathcal{M}_l$, write $\mathbf{b}_{jl}$ for an assignment to their common action variables.

In the upwards pass, $\mathcal{M}_j$ computes a conditional strategy for each assignment $\mathbf{b}_{jl}$ to its parent's actions. The value $Q_{jl}^{(t)}$ of this strategy is computed recursively:

$$Q_{jl}^{(t)}(\mathbf{b}_{jl}) = \max_{[\mathbf{b}_j \mid \mathbf{b}_{jl}]} \left[ Q_j^{(t)}(\mathbf{b}_j) + \sum_{k \in \text{ch}_j} Q_{kj}^{(t)}(\mathbf{b}_j) \right]$$

In the downwards pass, each subsystem chooses an action given the choices already made: $\arg\max_{[\mathbf{b}_j \mid \mathbf{b}_{jl}]} Q_{jl}^{(t)}(\mathbf{b}_j)$.

The cost of this action selection algorithm is linear in the number of subsystems and in the number of actions in each subsystem. Thus, we can efficiently compute the greedy policy associated with our compact value function.

## 9 Reusing subsystems, plans and messages

In typical real-world problems, subsystems of the same type will appear in several places in the subsystem tree. For example, in a car engine, there will typically be several cylinder subsystems. In addition to the conceptual advantages of representing all cylinders the same way, our algorithm can gain computational advantages by *reusing* both plans and messages in multiple parts of the subsystem tree.

We can view a subsystem tree (Definition 4.1) as a *class* or template. Then, when designing a factorization for a new problem, we can instantiate this class in multiple positions in our subsystem tree. We can also form complex classes out of simpler ones; instantiating a complex class then inserts a whole subtree into our tree (and also indicates how subsystems are grouped to form a hierarchical tree).

Now suppose that we have found a new policy for a subsystem $\mathcal{M}_j$. Our algorithm uses this policy to compute a set of dual variables $\phi$ as in (2), then marginalizes $\phi$ onto each of $\mathcal{M}_j$'s separating sets (8) to generate constraints in reward message LPs. These same dual variables $\phi$ are feasible for any subsystem $\mathcal{M}_k$ of the same class as $\mathcal{M}_j$.

Therefore, we can reuse $\phi$ by marginalizing it onto $\mathcal{M}_k$'s separating sets as well to generate extra constraints. Furthermore, we can record $\phi$ in $\mathcal{M}_j$'s class definition, and whenever a new subsystem tree uses another instance of $\mathcal{M}_j$'s class, we can save computation by reusing $\phi$ again.

Finally, if two whole subtrees of $\mathcal{M}$ are equivalent, we can reuse the subtree policy messages from our algorithm. More precisely, two subtrees are equivalent if their roots are of the same class and their children are equivalent. Sets of equivalent subtrees contain sets of same-class subsystems, and so policies from subsystems in one subtree can be reused in the others as described above. In addition, mixed policies for the whole subtree can be reused, since they will be feasible for one subtree iff they are feasible for the other. That means that whenever we add a row to $\Phi_j$ and $T_j$ (equations (9) and (10)) we can add the same row to $\Phi_k$ and $T_k$, yielding further computational savings.

## 10 Conclusions

In this paper, we presented a principled and practical planning algorithm for collaborative multiagent problems. We represent such problems using a hierarchical decomposition into local subsystems. Although each subsystem is small, once these subsystems are combined we can represent an exponentially larger problem.

Our planning algorithm can exploit this hierarchical structure for computational efficiency, avoiding an exponential blow-up. Furthermore, this algorithm can be implemented in a distributed fashion, where each agent only needs to solve local planning problems over its own subsystem. The global plan is computed by a message passing algorithm, where messages are calculated by local LPs.

Our representation and algorithm are suitable for heterogeneous systems, where subsystem MDPs are represented in different forms or solved by different algorithms. For example, one subsystem MDP could be solved by policy iteration, while other could be tackled with a library of heuristic policies. Furthermore, some subsystem MDPs could have known models, while others could be solved by reinforcement learning techniques.

Our planning algorithm is guaranteed to converge to the same solution as the centralized approach of Guestrin et al. [8], who experimentally tested the quality of their algorithm's policies on some benchmark problems. They concluded that the policies attained near-optimal performance on these problems and were significantly better than those produced by some other methods. Our distributed algorithm converges to the same policies; so, we would expect to see the same positive results, but with planning speedups from reuse and without the need for centralized planning.

We believe that hierarchical multiagent factored MDPs will facilitate the modeling of practical systems, while our distributed planning approach will make them applicable to the control of very large stochastic dynamical systems.

**Acknowledgments** We are very grateful to D. Koller, R. Parr, and M. Rosencrantz for many useful discussions. This work was supported by the DoD MURI, administered by the Office of Naval Research, Grant N00014-00-1-0637; Air Force contract F30602-00-2-0598, DARPA's TASK program; and AFRL contract F30602-01-C-0219, DARPA's MICA program. C. Guestrin was also supported by a Siebel Scholarship.